\documentclass[10pt, a4paper]{article}
\usepackage{lrec2022} % this is the new LREC2022 Style
\usepackage{multibib}
\newcites{languageresource}{Language Resources}
\usepackage{graphicx}
\usepackage{tabularx}
\usepackage{soul}
\usepackage{titlesec}
\titleformat{\section}{\normalfont\large\bfseries\center}{\thesection.}{1em}{}
\titleformat{\subsection}{\normalfont\SmallTitleFont\bfseries\raggedright}{\thesubsection.}{1em}{}
\titleformat{\subsubsection}{\normalfont\normalsize\bfseries\raggedright}{\thesubsubsection.}{1em}{}
\renewcommand\thesection{\arabic{section}}
\renewcommand\thesubsection{\thesection.\arabic{subsection}}
\renewcommand\thesubsubsection{\thesubsection.\arabic{subsubsection}}
\usepackage{algorithm}
\usepackage{algpseudocode}
\usepackage{listings}
\usepackage{epstopdf}
\usepackage[utf8]{inputenc}

\usepackage{hyperref}
\usepackage{xstring}
\usepackage{blindtext}

\usepackage{color}

\usepackage{bbding}

\algnewcommand\algorithmicforeach{\textbf{for each}}
\algdef{S}[FOR]{ForEach}[1]{\algorithmicforeach\ #1\ \algorithmicdo}

\title{DTW at Qur'an QA 2022: Utilising Transfer Learning with Transformers for Question Answering in a Low-resource Domain}

\name{Damith Premasiri\textsuperscript{1}, Tharindu Ranasinghe\textsuperscript{1}, Wajdi Zaghouani\textsuperscript{2}, Ruslan Mitkov\textsuperscript{1}} 

\address{\textsuperscript{1}University of Wolverhampton, UK \\ 
\textsuperscript{2}Hamad Bin Khalifa University, Qatar\\
         \{damith.premasiri, tharindu.ranasinghe, r.mitkov\}@wlv.ac.uk \\ wzaghouani@hbku.edu.qa\\}

\abstract{The task of machine reading comprehension (MRC) is a useful benchmark to evaluate the natural language understanding of machines. It has gained popularity in the natural language processing (NLP) field mainly due to the large number of datasets released for many languages. However, the research in MRC has been understudied in several domains, including religious texts. The goal of the Qur'an QA 2022 shared task is to fill this gap by producing state-of-the-art question answering and reading comprehension research on Qur'an. This paper describes the DTW entry to the Quran QA 2022 shared task. Our methodology uses transfer learning to take advantage of available Arabic MRC data. We further improve the results using various ensemble learning strategies. Our approach provided a partial Reciprocal Rank (pRR) score of 0.49 on the test set, proving its strong performance on the task.\\ \newline }

\begin{document}

\maketitleabstract

\section{Introduction}

Machine Reading Comprehension (MRC) is a challenging Natural Language Processing (NLP) application \cite{baradaran_ghiasi_amirkhani_2022}. The concept of MRC is similar to how humans are evaluated in examinations where a person should understand the text and answer questions based on the text. Similarly, the goal of a typical MRC task requires a machine to read a set of text passages and then answer questions about the passages. MRC systems could be widely applied in many NLP systems such as search engines and dialogue systems. Therefore, the NLP community has shown a great interest in MRC tasks over recent years.

The most common way of dealing with MRC tasks is to train a machine learning model on an annotated dataset. Over the years, researchers have experimented with different machine learning approaches ranging from traditional algorithms such as support vector machines \cite{suzuki-etal-2002-svm,YEN201377} to embedding based neural approaches such as transformers, with the latter providing state-of-the-art results in many datasets. We discuss them thoroughly in Section \ref{sec:related_work}. However, an annotated dataset is an essential requirement for these machine learning models. Identifying this, the NLP community has developed several datasets in recent years. The most popular MRC dataset is the Stanford Question Answering Dataset (SQuAD), which contains more than 100,000 annotated examples \cite{rajpurkar-etal-2016-squad}. SQuAD dataset has been extended to several languages including Arabic \cite{mozannar-etal-2019-neural}, Dutch \cite{10.1007/978-3-030-93842-0_9}, Persian \cite{9443126} and Sinhala \cite{10.1007/978-3-319-30933-0_32}. However, MRC datasets have been limited to common domains such as Wikipedia and MRC in low-resource domains, including religious books, have not been explored widely by the community \cite{baradaran_ghiasi_amirkhani_2022}. Moreover, most researchers focus on a few popular MRC datasets, while most other MRC datasets are not widely known and studied by the community \cite{app10217640}. 

Qur'an QA 2022 shared task\cite{malhas2022quranqa} has been organised to address these gaps in MRC research. The goal of the shared task is to trigger state-of-the-art question answering and reading comprehension research on a book that is sacredly held by more than 1.8 billion people across the world. The shared task relies on a recently released dataset of 1,337 question-passage-answer triplets extracted from the holy Qur'an \cite{10.1145/3400396}. Despite the novelty, the dataset poses several challenges. Firstly, since the dataset contains texts from Qur'an, modern embedding models would have problems encoding them. Therefore, we experimented with different pre-processing techniques to handle the texts from Qur'an. Secondly, the dataset is small compared to other MRC datasets such as SQuAD \cite{rajpurkar-etal-2016-squad}, and it would be difficult to fine-tune the state-of-the-art neural models. We experiment with different techniques such as transfer learning and ensemble learning to overcome this. We show that state-of-the-art neural models can be applied in smaller MRC datasets utilising the above methods. 

We address two research questions in this paper:

\vspace{2mm}

\textbf{RQ1:} Do ensemble models provide better results compared to single models?

\textbf{RQ2:} Can other Arabic MRC resources such as SOQAL~\cite{mozannar-etal-2019-neural} be used to improve the results for Qur’an MRC?

\vspace{2mm}

The code of the experiments has been released as an open-source Github project\footnote{The Github project is available on \url{https://github.com/DamithDR/QuestionAnswering}}. The project has been released as a Python package\footnote{The Python package is available on \url{https://pypi.org/project/quesans/}} and the pre-trained machine learning models are freely available to download in HuggingFace model hub\footnote{The pre-trained models are available on \url{https://huggingface.co/Damith/AraELECTRA-discriminator-SOQAL} and \url{https://huggingface.co/Damith/AraELECTRA-discriminator-QuranQA}}. Furthermore, we have created a docker image of the experiments adhering to the ACL reproducibility criteria\footnote{The docker image is available on \url{https://hub.docker.com/r/damithpremasiri/question-answering-quran}}. 

The rest of the paper is structured as follows. Section \ref{sec:related_work} presents an overview of MRC datasets and machine learning models. Section \ref{sec:data} describes the data we used in the experiments. In Section \ref{sec:method} we explain the experiments carried out. Section \ref{sec:results} discusses the results answering the research questions. Finally, the paper outlines future works and provides conclusions.

\section{Related Work}
\label{sec:related_work}
Machine reading comprehension is not newly proposed. The earliest known MRC system dates back to 1977 when \cite{lehnert1977process} developed a question answering program called the QUALM. In 1999 \cite{10.3115/1034678.1034731} constructed a reading comprehension system exploiting a corpus of 60 development and 60 test stories of 3rd to 6th-grade material. Due to the lack of high-quality MRC datasets and the poor performance of MRC models, this research field was understudied until the early 2010s. However, with the creation of large MRC datasets and with the success of word embedding based neural models in the NLP field, research in MRC has been popular in recent years. We present the related work in MRC in two broad categories; datasets and models. 

\paragraph{Datasets}
In 2013, \cite{richardson-etal-2013-mctest} created the MCTest dataset which contained 500 stories and 2000 questions. This dataset can be considered the first big MRC dataset. A breakthrough in MRC was achieved in 2015 when \cite{NIPS2015_afdec700} defined a new dataset generation method that provides large-scale supervised reading comprehension datasets. This was followed by the creation of large scale MRC datasets such as SQuAD\cite{rajpurkar-etal-2016-squad}. Later the SQuAD dataset has been expanded to many languages including Arabic \cite{mozannar-etal-2019-neural}, Dutch \cite{10.1007/978-3-030-93842-0_9}, French \cite{dhoffschmidt-etal-2020-fquad} and Russian \cite{10.1007/978-3-030-58219-7_1}. Furthermore, SQuAD has been extended to low-resource languages such as Persian \cite{9443126} and Sinhala \cite{10.1007/978-3-319-30933-0_32} proving that SQuAD has been an important benchmark in MRC research. MRC datasets have been compiled on different domains such as news \cite{trischler-etal-2017-newsqa}, publications \cite{dasigi-etal-2021-dataset} and natural sciences \cite{welbl-etal-2017-crowdsourcing}. As far as we know, Qur’an Reading Comprehension Dataset used in this shared task is the first dataset created on religious texts \cite{10.1145/3400396}.

\paragraph{Methods}
Most MRC systems in the early 2000s were rule-based or statistical models \cite{10.3115/1117595.1117598,10.3115/1117595.1117596}. These models do not provide good results compared to the neural methods introduced in recent years \cite{baradaran_ghiasi_amirkhani_2022}. \cite{NIPS2015_afdec700} developed a class of attention based deep neural networks that learn to read real documents and answer complex questions with minimal prior knowledge of language structure. Since 2015, with the emergence of various large scale, supervised datasets, neural network models have shown state-of-the-art results in MRC tasks. The recently introduced transformer models such as BERT \cite{devlin-etal-2019-bert} have already exceeded human performance over the related MRC benchmark datasets \cite{app10217640}. A critical contribution of the SQuAD benchmark is that it provides a system to submit the MRC models and a leaderboard to display the top results\footnote{SQuAD leaderboard is available on \url{https://rajpurkar.github.io/SQuAD-explorer/}}. This has enabled the NLP community to keep track of the state-of-the-art MRC systems. Other languages have also followed this approach\footnote{Korean MRC leaderboard is available on \url{https://korquad.github.io/}}. However, the NLP community has focused mainly on improving system performance on popular benchmarks such as SQuAD and has not focused on improving results on benchmarks with limited coverage, which we address in this research paper.

\section{Data}
\label{sec:data}
\begin{figure}[ht]
    \centering
    \includegraphics[scale=0.7]{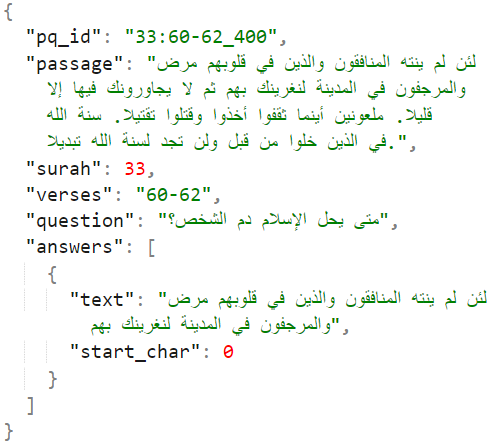}
    \caption{Sample Json object from the QRCD dataset \protect\cite{10.1145/3400396}}
    \label{fig:training_data_format}
\end{figure}

MRC tasks are usually divided into four categories: cloze style, multiple-choice, span prediction, and free form \cite{app9183698}. The Qur’an QA 2022 shared task\footnote{More information on the Qur’an QA 2022 shared task is available on \url{https://sites.google.com/view/quran-qa-2022/}} belongs to the span prediction category where the MRC system needs to select the correct beginning and end of the answer text from the context. The event organisers provided the QRCD (Quran Reading Comprehension Dataset), which contained 1,093 tuples of question-passage pairs that are coupled with their extracted answers to constitute 1,337 question-passage-answer triplets. QRCD is a JSON Lines (JSONL) file; each line is a JSON object that comprises a question-passage pair and its answers extracted from the accompanying passage. Figure \ref{fig:training_data_format} shows a sample training tuple. The distribution of the dataset into training, development and test sets is shown in Table \ref{tab:shared_task_data_table}. 

\begin{table}[h]
    \centering
    \scalebox{0.92}{
    \begin{tabular}{c|c|c|c}
        \hline
        \textbf{Dataset} & \textbf{\%} & \textbf{Q-P Pairs} & \textbf{Q-P-A Triplets} \\
        \hline
        Training & 65\% & 710 & 861 \\
        Development & 10\% & 109 & 128 \\
        Test & 25\% & 274 & 348 \\
        \hline
        All & 100\% & 1,093 & 1,337\\
        \hline
    \end{tabular}
    }
    \caption{Shared Task Data Composition. Column Q-P Pairs shows the number of Question Passage pairs, Column Q-P-A Triplets shows the number of Question Passage Answer triplets in the dataset}
    \label{tab:shared_task_data_table}
\end{table}

\paragraph{SOQAL} contains two Arabic MRC datasets; Arabic Reading Comprehension Dataset (ARCD) \cite{mozannar-etal-2019-neural}, composed of 1,395 questions posed by crowdworkers on Wikipedia articles, and a machine translation of the SQuAD \cite{mozannar-etal-2019-neural} containing  48,344 questions. SQuAD is widely used as the standard dataset in English MRC tasks, therefore using the machine translation of the same dataset will be helpful for the learning process. Compared to QRCD, SOQAL is a large dataset and both of these datasets belong to the span prediction MRC category. Therefore, they can be used to perform transfer learning which we describe in Section \ref{sec:method}. 

\section{Methodology}
\label{sec:method}
\begin{figure}[ht]
\centering
\includegraphics[scale=0.3]{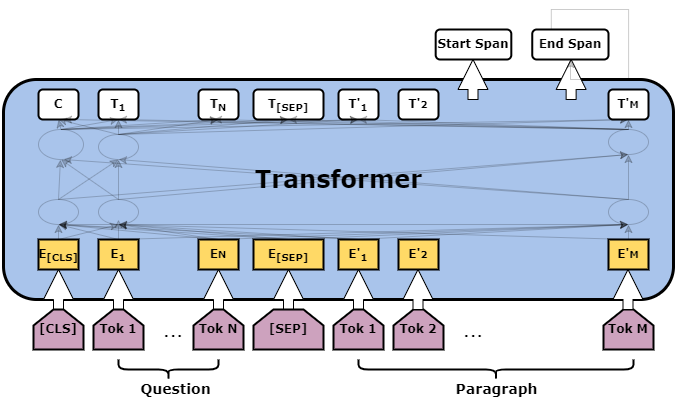}
\caption{Transformer Architecture for MRC}
\label{fig:architecture}
\end{figure}

With the introduction of BERT \cite{devlin-etal-2019-bert}, transformer models have achieved state-of-the-art results in different NLP applications such as text classification \cite{ranasinghe-hettiarachchi-2020-brums}, information extraction \cite{plum2022} and event detection \cite{giorgi-etal-2021-discovering}. Furthermore, the transformer architectures have shown promising results in SQuAD dataset \cite{Zhang_Yang_Zhao_2021,Zhang_Wu_Zhou_Duan_Zhao_Wang_2020,yamada-etal-2020-luke,Lan2020ALBERT}. In view of this, we use transformers as the basis of our methodology. Transformer architectures have been trained on general tasks like language modelling and then can be fine-tuned for MRC tasks. \cite{devlin-etal-2019-bert}. For the MRC task, transformer models take an input of a single sequence that contains the question and paragraph separated by a \textsc{[SEP]} token. Then the model introduces a start vector and an end vector. The probability of each word being the start-word is calculated by taking a dot product between the final embedding of the word and the start vector, followed by a softmax over all the words. The word with the highest probability value is considered. The architecture of transformer-based MRC model is shown in Figure \ref{fig:architecture}. 

We experimented with seven popular pre-trained transformer models that supports Arabic; camelbert-mix \cite{inoue-etal-2021-interplay}, camelbert-ca \cite{inoue-etal-2021-interplay}, mbert-cased \cite{devlin-etal-2019-bert} , mbert-uncased \cite{devlin-etal-2019-bert}, AraELECTRA-generator \cite{antoun-etal-2021-araelectra},  AraELECTRA-discriminator \cite{antoun-etal-2021-araelectra} and AraBERTv2 \cite{antoun-etal-2020-arabert}. These models are available in HuggingFace model hub \cite{wolf-etal-2020-transformers}. For all the experiments we used a batch-size of eight, Adam optimiser with learning rate $2\mathrm{e}{-5}$, and a linear learning rate warm-up over 10\% of the training data. During the training process, the parameters of the transformer model, as well as the parameters of the subsequent layers, were updated. The models were trained using only training data. All the models were trained for five epochs. For some of the experiments, we used an Nvidia GeForce RTX 2070 GPU, whilst for others we used a GeForce RTX 3090 GPU. This was purely based on the availability of the hardware and it was not a methodological decision.

We further used following fine-tuning strategies to improve the performance. 

\subsection{Ensemble Learning}
Ensemble learning is a popular technique in machine learning, where different machine learning models contribute to a single solution. As different machine learning algorithms tend to learn differently, the final predictions each one of them provides can be slightly different. However, they have the potential to contribute to the final output with ensemble learning. Usually, ensemble learning provides better results compared to single models \cite{https://doi.org/10.1002/widm.1249}.

Transformer models that we used as the base model are prone to the random seed \cite{hettiarachchi-ranasinghe-2020-infominer}. The same architecture can provide different results for different random seeds \cite{uyangodage-etal-2021-transformers}. To avoid the impact of this, we performed self ensemble. We trained the same architecture using five different random seeds and ensembled the output files using Algorithm \ref{alg:ensemble}. 

\begin{algorithm}
\caption{Ensemble Learning Algorithm for MRC}\label{alg:ensemble}
\begin{algorithmic}

\State $R \gets \textit{all results files}$
\State $r_i \gets \textit{i(th) result file}$
\State $Q \gets \textit{all questions}$
\State $q_j \gets \textit{j(th) question}$
\State $A \gets \textit{all answers}$
\State $a_{i,j} \gets \textit{answer for question j in files file i}$
\State $a_j \gets \textit{all unique answers for question j in all files}$
\State $a_{j,k} \gets \textit{answer k from unique answers for question j}$
\State $a_{i,j,m} \gets \textit{answer m from file i to question j}$
\State $answer_{j,k} \gets \textit{temporary score} $

\Repeat
    \ForEach{$a_{j,k} \in a_j$}
        \Repeat
            \ForEach {$a_{i,j,m} \in r_i $}
                \If{$a_{i,j,m} = a_{j,k}$}
                    \State $score_{j,k} \gets average(score_{a_{i,j,m}},score_{a_{j,k}})$
                    \State $answer_{j,k} \gets a_{j,k},score_{j,k}$
                \EndIf
            \EndFor
        \Until {$\textit{all items iterated in R}$}
        \State $\textit{final\_answers} \gets answer_{j,k}$
    \EndFor
\Until {$\textit{all unique answers iterated in for question j}$}

\State{$\textbf{answers} \gets \textbf{sort(answers)}$}

\Repeat
\ForEach {$q_j \in Q$}
    \Repeat
        \ForEach{$answer_{j,k} \in final_answers$}
            \State$rank_{j,k} \gets \textit{assign rank}$
        \EndFor
    \Until {$\textit{ iterate all answers for question j}$}
\EndFor
\Until {$\textit{iterate all questions iterated in Q}$}

\end{algorithmic}
\end{algorithm}

\subsection{Transfer Learning}
One limitation of the QRCD dataset is that training set only contains 710 annotated QnA pairs and as a result transformer models would find it difficult to properly fine-tune their weights in the training process. A common practice to overcome this is to utilise transfer learning.  The main idea of transfer learning is that we train a machine learning model on a resource rich setting, save the weights of the model and when we initialise the training process for a lower resource setting, start with the saved weights from the resource rich setting . Transfer learning has improved results for many NLP tasks such as offensive language identification \cite{ranasinghe-zampieri-2020-multilingual}, machine translation \cite{nguyen-chiang-2017-transfer} and named entity recognition \cite{lee-etal-2018-transfer}.

For this task, we first trained a transformer-based MRC model on SOQAL dataset which contained more training data compared to the QRCD dataset as mentioned in Section \ref{sec:data}. Then when we started the training for QRCD dataset we started from the saved weights from the SOQAL dataset.

\section{Results and Discussion}
\label{sec:results}
In this section, we report the experiments we conducted and their results. As advised by the task organisers, we used partial Reciprocal Rank (pRR) score to measure the model performance. It is a variant of the traditional Reciprocal Rank evaluation metric that considers partial matching. We also report Exact Match (EM), and F1@1 in the results tables, which are evaluation metrics applied only to the top predicted answer. The EM metric is a binary measure that rewards a system only if the top predicted answer matches exactly one of the gold answers. In comparison, the F1@1 metric measures the token overlap between the top predicted answer and the best matching gold answer. The reported results are for the dev set.

As can be seen in Table \ref{tab:benchmark_table}, camelbert-mix model produced the best results with 0.549 pRR value. This was closely followed by camelbert-ca and AraELECTRA-discriminator. Transformer models built specifically for Arabic generally outperformed multilingual models.

\begin{table}[ht]
    \centering
    \resizebox{7.5cm}{!}{
    \begin{tabular}{c|c|c|c}
         \textbf{Model} & \textbf{pRR} & \textbf{EM} & \textbf{F1@1} \\
         \hline
         AraELECTRA-discriminator & \textbf{0.516} & 0.303 & 0.495 \\
         AraELECTRA-generator & 0.355 & 0.339 & 0.324 \\
         camelbert-mix & \textbf{0.549} & 0.193 & 0.529 \\
         camelbert-ca & \textbf{0.535} & 0.119 & 0.516 \\
         mbert-cased & 0.425 & 0.321 & 0.405 \\
         mbert-uncased & 0.440 & 0.220 & 0.424 \\
         AraBERTv2 & 0.501 & 0.294 & 0.472 \\
    \end{tabular}
    }
    \caption{Results of different transformer models without ensemble learning or transfer-learning. Column \textbf{pRR} shows the partial Reciprocal Rank score, Column \textbf{EM} shows results for exact match and Column \textbf{F1@1} shows F1@1 score. The top three results are highlighted in Bold.}
    \label{tab:benchmark_table}
\end{table}

To answer our \textbf{RQ1}, we performed self ensemble learning. Table \ref{tab:ensembling_benchmark_table} shows the results of different models with results ensemble. Even though there was a slight improvement in AraELECTRA-discriminator, the overall impact for the results from the ensemble was very low. And we noticed that some of the models had performed less when using ensemble. However, the results were stable compared to the single models. Therefore, we used self ensemble learning even though it did not contribute to improving the results. With these findings, we answer our \textbf{RQ1}, ensemble models do not provide better results compared to single models; however, they provide more consistent results.

\begin{table}[h]
    \centering
    \resizebox{8cm}{!}{
        \begin{tabular}{c|c|c|c}
              \textbf{Model} & \textbf{pRR} & \textbf{EM} & \textbf{F1@1} \\
             \hline
             AraELECTRA-discriminator & \textbf{0.528} & 0.321 & 0.500 \\
             AraELECTRA-generator & 0.364 & 0.128 & 0.335 \\
             camelbert-mix & \textbf{0.520} & 0.303 & 0.497 \\
             camelbert-ca & \textbf{0.495} & 0.239 & 0.467 \\
             mbert-cased & 0.438 & 0.220 & 0.417 \\
             mbert-uncased & 0.424 & 0.220 & 0.399 \\
             AraBERTv2 & 0.475 & 0.239 & 0.436 \\
        \end{tabular}
    }
     \caption{Results of different transformer models with self ensemble learning. Column \textbf{pRR} shows the partial Reciprocal Rank score, Column \textbf{EM} shows results for exact match and Column \textbf{F1@1} shows F1@1 score. The top three results are highlighted in Bold.}
    \label{tab:ensembling_benchmark_table}
\end{table}

To answer our \textbf{RQ2}, we performed transfer learning from SOQAL \cite{mozannar-etal-2019-neural} to QRCD dataset as mentioned in Section \ref{sec:method}. We only conducted the experiments for the best model from the self ensemble setting. As can be seen in the results in Table \ref{tab:transfer_learned_table}, transfer learning improved the results for AraELECTRA-discriminator. Without transfer learning, AraELECTRA-discriminator scored only 0.528 pRR, while with transfer learning, it provided 0.616 pRR. We did not observe improvements in other transformer models. However, the 0.616 pRR we got with performing transfer learning with AraELECTRA-discriminator was the best result for the dev set. With this, we answer our \textbf{RQ2}, other Arabic MRC resources such as SOQAL~\cite{mozannar-etal-2019-neural} can be used to improve the results for Qur’an MRC. We believe that this finding will be important to the researchers working on low-resource MRC datasets.

\begin{table}[h]
    \centering
    \resizebox{7.5cm}{!}{
        \begin{tabular}{c|c|c|c}
              \textbf{Model} & \textbf{pRR} & \textbf{EM} & \textbf{F1@1} \\
             \hline
             AraELECTRA-discriminator & 0.616 & 0.394 & 0.609 \\
             camelbert-mix & 0.520 & 0.284 & 0.494 \\ 
             AraBERTv2 & 0.430 & 0.138 & 0.412 \\
        \end{tabular}
    }
    \caption{Results of different transformer models after transfer learning. Column \textbf{pRR} shows the partial Reciprocal Rank score, Column \textbf{EM} shows results for exact match and Column \textbf{F1@1} shows F1@1 score.}
    \label{tab:transfer_learned_table}
\end{table}

% The results in the Table \ref{tab:transfer_learned_table} are Transfer Learned models with ensemble results. Transfer Learning improved the performance of the models but for the development set, the improvement was not significant which was around 0.06 improvement in pRR. 

Based on the results of the dev set, we selected three models for the final submission; camelbert-mix with ensemble learning but without transfer learning, camelbert-mix with transfer learning and ensemble learning and AraELECTRA-discriminator with transfer learning and ensemble learning. Table \ref{tab:test_set_execution_results} shows the results that the organisers provided on the test set for our submitted models.

\begin{table}[h]
    \centering
    \resizebox{7.5cm}{!}{
        \begin{tabular}{c|c|c|c|c|c}
            \textbf{Model} & \textbf{TL} & \textbf{EN} & \textbf{pRR} & \textbf{EM} & \textbf{F1@1} \\
            \hline
            camelbert-mix & \XSolidBrush & \Checkmark & 0.290 & 0.084 & 0.258 \\
            camelbert-mix & \Checkmark &  \Checkmark & 0.408
            & 0.138 & 0.390 \\
            AraELECTRA-discriminator & \Checkmark & \Checkmark & \bf0.495 & 0.226 & 0.476
        \end{tabular}
    }
    \caption{Results of different transformer models on the test set. Column \textbf{TL} implies whether we performed transfer learning or not and the Column \textbf{EN} shows whether we performed ensemble learning. Column \textbf{pRR} shows the partial Reciprocal Rank score, Column \textbf{EM} shows results for exact match and Column \textbf{F1@1} shows F1@1 score.}
    \label{tab:test_set_execution_results}
\end{table}

AraELECTRA-discriminator performed best in the test set too. The camelbert-mix mode without transfer learning has decreased its performance from 0.549 to 0.290, which is a 47\% decrease. However, the models with transfer learning have performed comparatively high, confirming our answer to the \textbf{RQ2}. 

% Test results illustrates that the development set and the test set has significant differences which bert-base-arabic-camelbert-mix model without transfer learning has decreased its performance from 0.549 to 0.290 which is a 47 percent decrease of performance. However, the transfer learned models have performed comparatively high as these models are already having adjusted weights from a separate dataset.  

\section{Conclusion}
In this paper, we have presented the system submitted by the DTW team to the Qur'an QA 2022 shared task in the 5th Workshop on Open-Source Arabic Corpora and Processing Tools. We have shown that  AraELECTRA-discriminator with transfer learning from an Arabic MRC dataset is the most successful transformer model from several transformer models we experimented with. Our best system scored 0.495 pRR in the test set. With our \textbf{RQ1}, we showed that transformer models based on self ensemble provided stable results than single models in Qur'an QA task. Revisiting our \textbf{RQ2}, we showed that transfer learning could be used to improve the MRC results of the Qur'an. We believe that this finding would pave the way to enhance MRC in many low-resource domains. Our code, software and the pre-trained models have been made available freely to the researchers working on similar problems.

In future work, we would like to explore more to transfer learning. We will be exploring cross-lingual transfer learning with larger English MRC datasets such as SQuAD, as cross-lingual transfer learning has shown splendid results in many NLP tasks \cite{ranasinghe-etal-2021-exploratory}. Furthermore we will be exploring zero-shot and few-shot learning, which could benefit a multitude of low-resource languages.

\section{Acknowledgements}

This project was partially funded by the University of Wolverhampton's RIF4 Research Investment Funding provided for the Responsible Digital Humanities lab (RIGHT).

We would like to thank the Qur'an QA 2022 shared task organisers for running this interesting shared task and for replying promptly to all our inquiries. Furthermore, we thank the anonymous OSACT 2022 reviewers for their insightful feedback.

% \nocite{*}
\section{References}\label{reference}
%\label{main:ref}

\bibliographystyle{lrec2022-bib}
\bibliography{lrec2022-example}

\begin{thebibliography}{}

\bibitem[\protect\citename{Abadani \bgroup et al.\egroup }2021]{9443126}
Abadani, N., Mozafari, J., Fatemi, A., Nematbakhsh, M.~A., and Kazemi, A.
\newblock (2021).
\newblock Parsquad: Machine translated squad dataset for persian question
  answering.
\newblock In {\em 2021 7th International Conference on Web Research (ICWR)},
  pages 163--168.

\bibitem[\protect\citename{Antoun \bgroup et al.\egroup
  }2020]{antoun-etal-2020-arabert}
Antoun, W., Baly, F., and Hajj, H.
\newblock (2020).
\newblock {A}ra{BERT}: Transformer-based model for {A}rabic language
  understanding.
\newblock In {\em Proceedings of the 4th Workshop on Open-Source Arabic Corpora
  and Processing Tools, with a Shared Task on Offensive Language Detection},
  pages 9--15, Marseille, France, May. European Language Resource Association.

\bibitem[\protect\citename{Antoun \bgroup et al.\egroup
  }2021]{antoun-etal-2021-araelectra}
Antoun, W., Baly, F., and Hajj, H.
\newblock (2021).
\newblock {A}ra{ELECTRA}: Pre-training text discriminators for {A}rabic
  language understanding.
\newblock In {\em Proceedings of the Sixth Arabic Natural Language Processing
  Workshop}, pages 191--195, Kyiv, Ukraine (Virtual), April. Association for
  Computational Linguistics.

\bibitem[\protect\citename{Baradaran \bgroup et al.\egroup
  }2022]{baradaran_ghiasi_amirkhani_2022}
Baradaran, R., Ghiasi, R., and Amirkhani, H.
\newblock (2022).
\newblock A survey on machine reading comprehension systems.
\newblock {\em Natural Language Engineering}, page 1–50.

\bibitem[\protect\citename{Charniak \bgroup et al.\egroup
  }2000]{10.3115/1117595.1117596}
Charniak, E., Altun, Y., Braz, R. d.~S., Garrett, B., Kosmala, M., Moscovich,
  T., Pang, L., Pyo, C., Sun, Y., Wy, W., Yang, Z., Zeller, S., and Zorn, L.
\newblock (2000).
\newblock Reading comprehension programs in a statistical-language-processing
  class.
\newblock In {\em Proceedings of the 2000 ANLP/NAACL Workshop on Reading
  Comprehension Tests as Evaluation for Computer-Based Language Understanding
  Sytems - Volume 6}, ANLP/NAACL-ReadingComp '00, page 1–5, USA. Association
  for Computational Linguistics.

\bibitem[\protect\citename{Dasigi \bgroup et al.\egroup
  }2021]{dasigi-etal-2021-dataset}
Dasigi, P., Lo, K., Beltagy, I., Cohan, A., Smith, N.~A., and Gardner, M.
\newblock (2021).
\newblock A dataset of information-seeking questions and answers anchored in
  research papers.
\newblock In {\em Proceedings of the 2021 Conference of the North American
  Chapter of the Association for Computational Linguistics: Human Language
  Technologies}, pages 4599--4610, Online, June. Association for Computational
  Linguistics.

\bibitem[\protect\citename{Devlin \bgroup et al.\egroup
  }2019]{devlin-etal-2019-bert}
Devlin, J., Chang, M.-W., Lee, K., and Toutanova, K.
\newblock (2019).
\newblock {BERT}: Pre-training of deep bidirectional transformers for language
  understanding.
\newblock In {\em Proceedings of the 2019 Conference of the North {A}merican
  Chapter of the Association for Computational Linguistics: Human Language
  Technologies, Volume 1 (Long and Short Papers)}, pages 4171--4186,
  Minneapolis, Minnesota, June. Association for Computational Linguistics.

\bibitem[\protect\citename{d{'}Hoffschmidt \bgroup et al.\egroup
  }2020]{dhoffschmidt-etal-2020-fquad}
d{'}Hoffschmidt, M., Belblidia, W., Heinrich, Q., Brendl{\'e}, T., and Vidal,
  M.
\newblock (2020).
\newblock {FQ}u{AD}: {F}rench question answering dataset.
\newblock In {\em Findings of the Association for Computational Linguistics:
  EMNLP 2020}, pages 1193--1208, Online, November. Association for
  Computational Linguistics.

\bibitem[\protect\citename{Efimov \bgroup et al.\egroup
  }2020]{10.1007/978-3-030-58219-7_1}
Efimov, P., Chertok, A., Boytsov, L., and Braslavski, P.
\newblock (2020).
\newblock Sberquad -- russian reading comprehension dataset: Description and
  analysis.
\newblock In Avi Arampatzis, et~al., editors, {\em Experimental IR Meets
  Multilinguality, Multimodality, and Interaction}, pages 3--15, Cham. Springer
  International Publishing.

\bibitem[\protect\citename{Giorgi \bgroup et al.\egroup
  }2021]{giorgi-etal-2021-discovering}
Giorgi, S., Zavarella, V., Tanev, H., Stefanovitch, N., Hwang, S.,
  Hettiarachchi, H., Ranasinghe, T., Kalyan, V., Tan, P., Tan, S., Andrews, M.,
  Hu, T., Stoehr, N., Re, F.~I., Vegh, D., Atzenhofer, D., Curtis, B., and
  H{\"u}rriyeto{\u{g}}lu, A.
\newblock (2021).
\newblock Discovering black lives matter events in the {U}nited {S}tates:
  Shared task 3, {CASE} 2021.
\newblock In {\em Proceedings of the 4th Workshop on Challenges and
  Applications of Automated Extraction of Socio-political Events from Text
  (CASE 2021)}, pages 218--227, Online, August. Association for Computational
  Linguistics.

\bibitem[\protect\citename{Hermann \bgroup et al.\egroup
  }2015]{NIPS2015_afdec700}
Hermann, K.~M., Kocisky, T., Grefenstette, E., Espeholt, L., Kay, W., Suleyman,
  M., and Blunsom, P.
\newblock (2015).
\newblock Teaching machines to read and comprehend.
\newblock In C.~Cortes, et~al., editors, {\em Advances in Neural Information
  Processing Systems}, volume~28. Curran Associates, Inc.

\bibitem[\protect\citename{Hettiarachchi and
  Ranasinghe}2020]{hettiarachchi-ranasinghe-2020-infominer}
Hettiarachchi, H. and Ranasinghe, T.
\newblock (2020).
\newblock {I}nfo{M}iner at {WNUT}-2020 task 2: Transformer-based covid-19
  informative tweet extraction.
\newblock In {\em Proceedings of the Sixth Workshop on Noisy User-generated
  Text (W-NUT 2020)}, pages 359--365, Online, November. Association for
  Computational Linguistics.

\bibitem[\protect\citename{Hirschman \bgroup et al.\egroup
  }1999]{10.3115/1034678.1034731}
Hirschman, L., Light, M., Breck, E., and Burger, J.~D.
\newblock (1999).
\newblock Deep read: A reading comprehension system.
\newblock ACL '99, page 325–332, USA. Association for Computational
  Linguistics.

\bibitem[\protect\citename{Inoue \bgroup et al.\egroup
  }2021]{inoue-etal-2021-interplay}
Inoue, G., Alhafni, B., Baimukan, N., Bouamor, H., and Habash, N.
\newblock (2021).
\newblock The interplay of variant, size, and task type in {A}rabic pre-trained
  language models.
\newblock In {\em Proceedings of the Sixth Arabic Natural Language Processing
  Workshop}, pages 92--104, Kyiv, Ukraine (Virtual), April. Association for
  Computational Linguistics.

\bibitem[\protect\citename{Jayakody \bgroup et al.\egroup
  }2016]{10.1007/978-3-319-30933-0_32}
Jayakody, J. A. T.~K., Gamlath, T. S.~K., Lasantha, W. A.~N., Premachandra, K.
  M. K.~P., Nugaliyadde, A., and Mallawarachchi, Y.
\newblock (2016).
\newblock ``mahoshadha'', the sinhala tagged corpus based question answering
  system.
\newblock In Suresh~Chandra Satapathy et~al., editors, {\em Proceedings of
  First International Conference on Information and Communication Technology
  for Intelligent Systems: Volume 1}, pages 313--322, Cham. Springer
  International Publishing.

\bibitem[\protect\citename{Lan \bgroup et al.\egroup }2020]{Lan2020ALBERT}
Lan, Z., Chen, M., Goodman, S., Gimpel, K., Sharma, P., and Soricut, R.
\newblock (2020).
\newblock Albert: A lite bert for self-supervised learning of language
  representations.
\newblock In {\em International Conference on Learning Representations}.

\bibitem[\protect\citename{Lee \bgroup et al.\egroup
  }2018]{lee-etal-2018-transfer}
Lee, J.~Y., Dernoncourt, F., and Szolovits, P.
\newblock (2018).
\newblock Transfer learning for named-entity recognition with neural networks.
\newblock In {\em Proceedings of the Eleventh International Conference on
  Language Resources and Evaluation ({LREC} 2018)}, Miyazaki, Japan, May.
  European Language Resources Association (ELRA).

\bibitem[\protect\citename{Lehnert}1977]{lehnert1977process}
Lehnert, W.~G.
\newblock (1977).
\newblock {\em The process of question answering.}
\newblock Yale University.

\bibitem[\protect\citename{Liu \bgroup et al.\egroup }2019]{app9183698}
Liu, S., Zhang, X., Zhang, S., Wang, H., and Zhang, W.
\newblock (2019).
\newblock Neural machine reading comprehension: Methods and trends.
\newblock {\em Applied Sciences}, 9(18).

\bibitem[\protect\citename{Malhas and Elsayed}2020]{10.1145/3400396}
Malhas, R. and Elsayed, T.
\newblock (2020).
\newblock Ayatec: Building a reusable verse-based test collection for arabic
  question answering on the holy qur’an.
\newblock {\em ACM Trans. Asian Low-Resour. Lang. Inf. Process.}, 19(6), oct.

\bibitem[\protect\citename{Malhas \bgroup et al.\egroup
  }2022]{malhas2022quranqa}
Malhas, R., Mansour, W., and Elsayed, T.
\newblock (2022).
\newblock Qur'an {QA} 2022: Overview of the first shared task on question
  answering over the holy qur'an.
\newblock In {\em Proceedings of the 5th Workshop on Open-Source Arabic Corpora
  and Processing Tools (OSACT5) at the 13th Language Resources and Evaluation
  Conference (LREC 2022)}.

\bibitem[\protect\citename{Mozannar \bgroup et al.\egroup
  }2019]{mozannar-etal-2019-neural}
Mozannar, H., Maamary, E., El~Hajal, K., and Hajj, H.
\newblock (2019).
\newblock Neural {A}rabic question answering.
\newblock In {\em Proceedings of the Fourth Arabic Natural Language Processing
  Workshop}, pages 108--118, Florence, Italy, August. Association for
  Computational Linguistics.

\bibitem[\protect\citename{Nguyen and Chiang}2017]{nguyen-chiang-2017-transfer}
Nguyen, T.~Q. and Chiang, D.
\newblock (2017).
\newblock Transfer learning across low-resource, related languages for neural
  machine translation.
\newblock In {\em Proceedings of the Eighth International Joint Conference on
  Natural Language Processing (Volume 2: Short Papers)}, pages 296--301,
  Taipei, Taiwan, November. Asian Federation of Natural Language Processing.

\bibitem[\protect\citename{Plum \bgroup et al.\egroup }2022]{plum2022}
Plum, A., Ranasinghe, T., Jones, S., Orasan, C., and Mitkov, R.
\newblock (2022).
\newblock Biographical: A semi-supervised relation extraction dataset.
\newblock In {\em Proceedings of the 45th International ACM SIGIR Conference on
  Research and Development in Information Retrieval}, Madrid, Spain.
  Association for Computing Machinery.

\bibitem[\protect\citename{Rajpurkar \bgroup et al.\egroup
  }2016]{rajpurkar-etal-2016-squad}
Rajpurkar, P., Zhang, J., Lopyrev, K., and Liang, P.
\newblock (2016).
\newblock {SQ}u{AD}: 100,000+ questions for machine comprehension of text.
\newblock In {\em Proceedings of the 2016 Conference on Empirical Methods in
  Natural Language Processing}, pages 2383--2392, Austin, Texas, November.
  Association for Computational Linguistics.

\bibitem[\protect\citename{Ranasinghe and
  Hettiarachchi}2020]{ranasinghe-hettiarachchi-2020-brums}
Ranasinghe, T. and Hettiarachchi, H.
\newblock (2020).
\newblock {BRUMS} at {S}em{E}val-2020 task 12: Transformer based multilingual
  offensive language identification in social media.
\newblock In {\em Proceedings of the Fourteenth Workshop on Semantic
  Evaluation}, pages 1906--1915, Barcelona (online), December. International
  Committee for Computational Linguistics.

\bibitem[\protect\citename{Ranasinghe and
  Zampieri}2020]{ranasinghe-zampieri-2020-multilingual}
Ranasinghe, T. and Zampieri, M.
\newblock (2020).
\newblock Multilingual offensive language identification with cross-lingual
  embeddings.
\newblock In {\em Proceedings of the 2020 Conference on Empirical Methods in
  Natural Language Processing (EMNLP)}, pages 5838--5844, Online, November.
  Association for Computational Linguistics.

\bibitem[\protect\citename{Ranasinghe \bgroup et al.\egroup
  }2021]{ranasinghe-etal-2021-exploratory}
Ranasinghe, T., Orasan, C., and Mitkov, R.
\newblock (2021).
\newblock An exploratory analysis of multilingual word-level quality estimation
  with cross-lingual transformers.
\newblock In {\em Proceedings of the 59th Annual Meeting of the Association for
  Computational Linguistics and the 11th International Joint Conference on
  Natural Language Processing (Volume 2: Short Papers)}, pages 434--440,
  Online, August. Association for Computational Linguistics.

\bibitem[\protect\citename{Richardson \bgroup et al.\egroup
  }2013]{richardson-etal-2013-mctest}
Richardson, M., Burges, C.~J., and Renshaw, E.
\newblock (2013).
\newblock {MCT}est: A challenge dataset for the open-domain machine
  comprehension of text.
\newblock In {\em Proceedings of the 2013 Conference on Empirical Methods in
  Natural Language Processing}, pages 193--203, Seattle, Washington, USA,
  October. Association for Computational Linguistics.

\bibitem[\protect\citename{Riloff and Thelen}2000]{10.3115/1117595.1117598}
Riloff, E. and Thelen, M.
\newblock (2000).
\newblock A rule-based question answering system for reading comprehension
  tests.
\newblock In {\em Proceedings of the 2000 ANLP/NAACL Workshop on Reading
  Comprehension Tests as Evaluation for Computer-Based Language Understanding
  Sytems - Volume 6}, ANLP/NAACL-ReadingComp '00, page 13–19, USA.
  Association for Computational Linguistics.

\bibitem[\protect\citename{Rouws \bgroup et al.\egroup
  }2022]{10.1007/978-3-030-93842-0_9}
Rouws, N.~J., Vakulenko, S., and Katrenko, S.
\newblock (2022).
\newblock Dutch squad and ensemble learning for question answering
  from labour agreements.
\newblock In Luis~A. Leiva, et~al., editors, {\em Artificial Intelligence and
  Machine Learning}, pages 155--169, Cham. Springer International Publishing.

\bibitem[\protect\citename{Sagi and
  Rokach}2018]{https://doi.org/10.1002/widm.1249}
Sagi, O. and Rokach, L.
\newblock (2018).
\newblock Ensemble learning: A survey.
\newblock {\em WIREs Data Mining and Knowledge Discovery}, 8(4):e1249.

\bibitem[\protect\citename{Suzuki \bgroup et al.\egroup
  }2002]{suzuki-etal-2002-svm}
Suzuki, J., Sasaki, Y., and Maeda, E.
\newblock (2002).
\newblock {SVM} answer selection for open-domain question answering.
\newblock In {\em {COLING} 2002: The 19th International Conference on
  Computational Linguistics}.

\bibitem[\protect\citename{Trischler \bgroup et al.\egroup
  }2017]{trischler-etal-2017-newsqa}
Trischler, A., Wang, T., Yuan, X., Harris, J., Sordoni, A., Bachman, P., and
  Suleman, K.
\newblock (2017).
\newblock {N}ews{QA}: A machine comprehension dataset.
\newblock In {\em Proceedings of the 2nd Workshop on Representation Learning
  for {NLP}}, pages 191--200, Vancouver, Canada, August. Association for
  Computational Linguistics.

\bibitem[\protect\citename{Uyangodage \bgroup et al.\egroup
  }2021]{uyangodage-etal-2021-transformers}
Uyangodage, L., Ranasinghe, T., and Hettiarachchi, H.
\newblock (2021).
\newblock Transformers to fight the {COVID}-19 infodemic.
\newblock In {\em Proceedings of the Fourth Workshop on NLP for Internet
  Freedom: Censorship, Disinformation, and Propaganda}, pages 130--135, Online,
  June. Association for Computational Linguistics.

\bibitem[\protect\citename{Welbl \bgroup et al.\egroup
  }2017]{welbl-etal-2017-crowdsourcing}
Welbl, J., Liu, N.~F., and Gardner, M.
\newblock (2017).
\newblock Crowdsourcing multiple choice science questions.
\newblock In {\em Proceedings of the 3rd Workshop on Noisy User-generated
  Text}, pages 94--106, Copenhagen, Denmark, September. Association for
  Computational Linguistics.

\bibitem[\protect\citename{Wolf \bgroup et al.\egroup
  }2020]{wolf-etal-2020-transformers}
Wolf, T., Debut, L., Sanh, V., Chaumond, J., Delangue, C., Moi, A., Cistac, P.,
  Rault, T., Louf, R., Funtowicz, M., Davison, J., Shleifer, S., von Platen,
  P., Ma, C., Jernite, Y., Plu, J., Xu, C., Le~Scao, T., Gugger, S., Drame, M.,
  Lhoest, Q., and Rush, A.
\newblock (2020).
\newblock Transformers: State-of-the-art natural language processing.
\newblock In {\em Proceedings of the 2020 Conference on Empirical Methods in
  Natural Language Processing: System Demonstrations}, pages 38--45, Online,
  October. Association for Computational Linguistics.

\bibitem[\protect\citename{Yamada \bgroup et al.\egroup
  }2020]{yamada-etal-2020-luke}
Yamada, I., Asai, A., Shindo, H., Takeda, H., and Matsumoto, Y.
\newblock (2020).
\newblock {LUKE}: Deep contextualized entity representations with entity-aware
  self-attention.
\newblock In {\em Proceedings of the 2020 Conference on Empirical Methods in
  Natural Language Processing (EMNLP)}, pages 6442--6454, Online, November.
  Association for Computational Linguistics.

\bibitem[\protect\citename{Yen \bgroup et al.\egroup }2013]{YEN201377}
Yen, S.-J., Wu, Y.-C., Yang, J.-C., Lee, Y.-S., Lee, C.-J., and Liu, J.-J.
\newblock (2013).
\newblock A support vector machine-based context-ranking model for question
  answering.
\newblock {\em Information Sciences}, 224:77--87.

\bibitem[\protect\citename{Zeng \bgroup et al.\egroup }2020]{app10217640}
Zeng, C., Li, S., Li, Q., Hu, J., and Hu, J.
\newblock (2020).
\newblock A survey on machine reading comprehension—tasks, evaluation metrics
  and benchmark datasets.
\newblock {\em Applied Sciences}, 10(21).

\bibitem[\protect\citename{Zhang \bgroup et al.\egroup
  }2020]{Zhang_Wu_Zhou_Duan_Zhao_Wang_2020}
Zhang, Z., Wu, Y., Zhou, J., Duan, S., Zhao, H., and Wang, R.
\newblock (2020).
\newblock Sg-net: Syntax-guided machine reading comprehension.
\newblock {\em Proceedings of the AAAI Conference on Artificial Intelligence},
  34(05):9636--9643, Apr.

\bibitem[\protect\citename{Zhang \bgroup et al.\egroup
  }2021]{Zhang_Yang_Zhao_2021}
Zhang, Z., Yang, J., and Zhao, H.
\newblock (2021).
\newblock Retrospective reader for machine reading comprehension.
\newblock {\em Proceedings of the AAAI Conference on Artificial Intelligence},
  35(16):14506--14514, May.

\end{thebibliography}

% \section{Language Resource References}
% \label{lr:ref}
% \bibliographystylelanguageresource{lrec2022-bib}
% \bibliographylanguageresource{languageresource}

\end{document}